%% file: acl_latex.tex
\pdfoutput=1

\documentclass[11pt]{article}

\usepackage[final]{acl}

\usepackage{times}
\usepackage{latexsym}
\usepackage{amsmath}
\usepackage{graphicx}

\usepackage[T1]{fontenc}

\usepackage[utf8]{inputenc}

\usepackage{microtype}

\usepackage{inconsolata}

\usepackage{graphicx}
\usepackage{amssymb}
\usepackage{pifont}
\title{Negative Matters: Multi-Granularity Hard-Negative Synthesis and Anchor-Token-Aware Pooling for Enhanced Text Embeddings}

\author{
 \textbf{Tengyu Pan\textsuperscript{1}},
 \textbf{Zhichao Duan\textsuperscript{1}},
 \textbf{Zhenyu Li\textsuperscript{1}},
 \textbf{Bowen Dong\textsuperscript{1}},
\\
 \textbf{Ning Liu\textsuperscript{2}},
 \textbf{Xiuxing Li\textsuperscript{3}},
 \textbf{Jianyong Wang\textsuperscript{1}\thanks{indicates corresponding author.}}
\\
\\
\textsuperscript{1}{Tsinghua University, Beijing, China.}
\textsuperscript{2}{Shandong University, Shandong, China.}
\\
\textsuperscript{3}{Beijing Institute of Technology, Beijing, China.}
\\
{\tt pty23@mails.tsinghua.edu.cn}
}

\begin{document}
\maketitle
\begin{abstract}

Text embedding models are essential for various natural language processing tasks, enabling the effective encoding of semantic information into dense vector representations. These models are typically optimized using triplets of (query, positive, negative) data pairs for contrastive learning, where the negative samples play a critical role in enhancing the model's ability to discern subtle semantic distinctions. In this work, we introduce a \textbf{M}ulti-\textbf{G}ranularity \textbf{H}ard-negative (MGH) synthesis framework that leverages large language models (LLMs) to generate diverse negative samples with varying levels of similarity with the query. This approach facilitates a coarse-to-fine curriculum learning strategy during supervised training, allowing the embedding model to progressively learn more nuanced semantic representations. Meanwhile, we propose an \textbf{A}nchor \textbf{T}oken \textbf{A}ware (ATA) pooling method that assigns higher weights to anchor tokens based on aggregation patterns observed in LLMs, improving text embedding accuracy  without increasing model complexity.  Comprehensive experiments on the MTEB benchmark demonstrate that our methods achieve state-of-the-art performance, surpassing existing synthesis strategies both with synthetic data and when combined with public retrieval datasets.

\end{abstract}

\section{Introduction}

\begin{figure}[t]
  \includegraphics[width=\columnwidth]{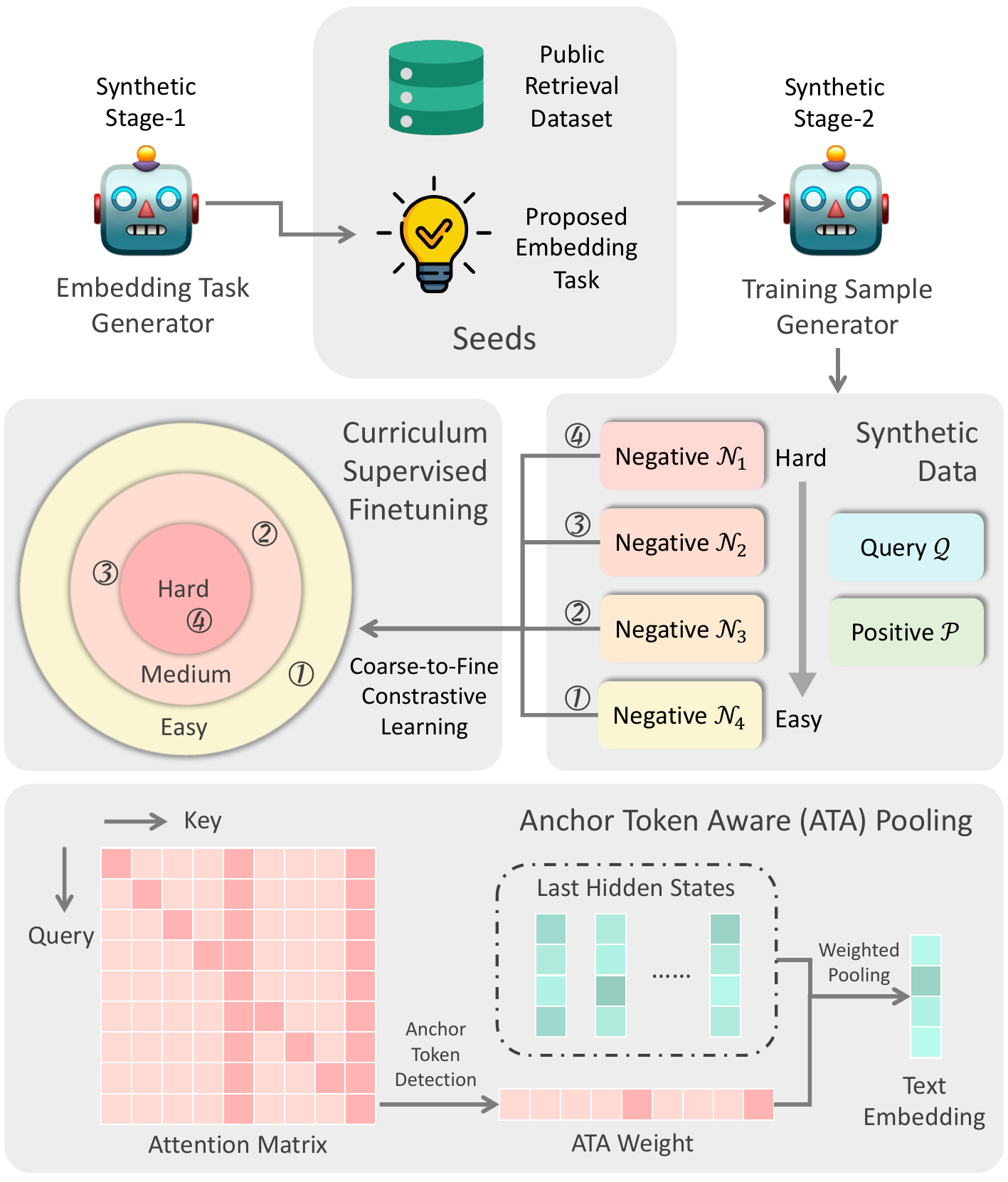}
  \caption{Illustration of our proposed multi-granularity hard-negative sample generation and coarse-to-fine learning paradigm. \ding{172}\textasciitilde \ding{175} indicate the training order of the synthetic negative samples.}
  \label{fig:framework}
\end{figure}

Text embedding models are designed to encode the semantic meaning of a given sequence of natural language words, sentences, or larger text spans into dense vector representations.  These vector representations capture not only the lexical content of the text but also its syntactic and semantic nuances, facilitating a wide range of downstream natural language processing (NLP) tasks such as sentiment analysis, text clustering, and content-based information retrieval.

Previous studies have explored the potential of leveraging large language models (LLMs) to generate synthetic data for text embedding training tasks \citep{Bonifacio2022InParsDA,wang2024improving,lee2024gecko}. The substantial volume of synthetic data contributes to increased diversity, thereby improving the model's robustness across various encoding tasks. However, generating high-quality hard negative samples required by contrastive learning is a challenging task for synthetic models, as an effective hard negative must maintain the right balance in its distinction from the positive examples. If the hard negative is overly similar to the query, it may confuse the model with positive samples, whereas if there is a significant difference in the content, the model may struggle to extract useful information from hard negative samples.

In this light, we propose a data synthesis framework to fully leverage the LLMs' ability to identify the partial order of text similarity, facilitating the generation of multi-granularity synthetic data. By simultaneously generating hard negative samples at multiple similarity levels, the synthesized data not only improves overall quality but also allows for controlled difficulty of the negatives. In the subsequent supervised training, we use the generated difficulty-controllable data to implement coarse-to-fine curriculum learning. By progressively moving from simple to hard training samples, the embedding model learns increasingly complex representations, resulting in improved stability and effectiveness.

Moreover, as research increasingly focuses on effectively adapting large models into text embedding models, studies \citep{lee2024nv, behnamghader2024llm2vec} have explored converting causal attention to bidirectional attention to enhance embedding performance. In this context, two common methods to obtain embeddings from the hidden states of a sequence of tokens are \textit{mean pooling}, which averages the final hidden states, and \textit{last token pooling}, which uses the last hidden state of the <EOS> token as the sentence representation vector. However, mean pooling tends to dilute the critical information tokens when averaging across all tokens, which leads to a loss of significant features \citep{lee2024nv}. In contrast, the last token pooling method is sensitive to noisy information within the sentence, resulting in instability in the encoding \citep{springer2024repetition}. Recently, NV-Embed \citep{lee2024nv} addressed the insufficient pooling issue by adding a cross-attention layer over the tokens' final hidden states in a dictionary learning method.

However, we suggest that a simple yet effective pooling method can be utilized without introducing additional parameters. Recent studies have revealed the aggregation pattern of large language models \citep{wang2023label,huang2024opera}, indicating that decoder-only models tend to aggregate textual information into anchor tokens at shallow layers and use these tokens to generate the next token in deeper layers. 

In this paper, we observe that the aggregation pattern still holds in models transformed from causal to bidirectional attention. By simply assigning greater weight to anchor tokens, we achieve improved accuracy in text embedding tasks compared to conventional pooling methods.

Our contributions are summarized as follows:

1. \textbf{A MGH Data Synthesis Framework}. We propose a \textbf{M}ulti-\textbf{G}ranularity \textbf{H}ard-negative framework that effectively generates diverse negative samples of varying difficulty levels, fully leveraging the large model's capability to discern the partial order of text similarity. The framework allows for controlled progression in the difficulty of the generated negative examples, enabling subsequent text embedding models to learn a more accurate embedding representation through a coarse-to-fine manner.

2. \textbf{An ATA Pooling Method}. We propose an \textbf{A}nchor \textbf{T}oken \textbf{A}ware pooling method that effectively leverages the aggregation pattern of LLMs to acquire a more accurate sentence representation. The model trained with ATA pooling outperformed previous pooling methods and, when trained solely on publicly available retrieval data without MTEB training split, achieved the state-of-the-art model on the MTEB leaderboard.

\section{Method}

\subsection{Multi-granularity Synthetic Data Generation}

The overall framework of our proposed data synthesis method is illustrated in Figure \ref{fig:framework}, which consists of two primary stages. In accordance with the setup of \citet{wang2024improving}, we begin by querying LLMs to generate a list of potential tasks, categorized into two types: (1) asymmetric text matching tasks comprising four subtasks including short-long match, long-short match, long-long match, and short-short match; (2) symmetric task represented by semantic textual similarity (STS). The task brainstorming process in stage 1 generates a wide range of embedding tasks to enrich the diversity of synthetic data. Further details of the task types are provided in Appendix \ref{sec:appendix_details_of_categories}.

In stage 2, using a diverse set of tasks as seeds, we query the LLM to generate \texttt{(query, positive, negative)} samples for subsequent contrastive supervised training, which are then used in contrastive supervised training with a standard infoNCE objective to minimize the distance between query and positive samples, while maximizing the separation from negative samples:

\begin{equation}
  \label{eq:info_nce}
 \mathcal{L} =  -\log\frac{e^{\tau\cdot\phi(q,d^+)}}{e^{\tau\cdot\phi(q,d^+)} + \sum_{d^- \in N}{e^{\tau\cdot\phi(q,d^-)}}}
\end{equation}

where $\phi$ denotes the cosine similarity function, $\tau$ is the temperature parameter and is set to 1. Here, $q$, $d^+$ and $d^-$ represent the query, positive, and negative samples respectively. Recognizing that the quality of negative samples significantly impacts the supervised training of text-embedding models, we aim to fully leverage the large model's ability to distinguish between different granularities of negative samples during the generation process. To be specific, we formulate the synthetic target as follows:

\begin{equation}
\mathcal{T}^S = \{(\mathcal{Q}^S, \mathcal{P}^S, \{\mathcal{N}^S_k\})\}
\end{equation}

where  $\mathcal{Q}^S$ represents an example query,  $\mathcal{P}^S$  denotes its corresponding positive sample, and $\{\mathcal{N}^S_k\}$  refers to a set of hard negative samples with varying levels of granularity indexed by $k$, which is set to 4 in the following experiments.

\begin{figure}[t]
  \includegraphics[width=\columnwidth]{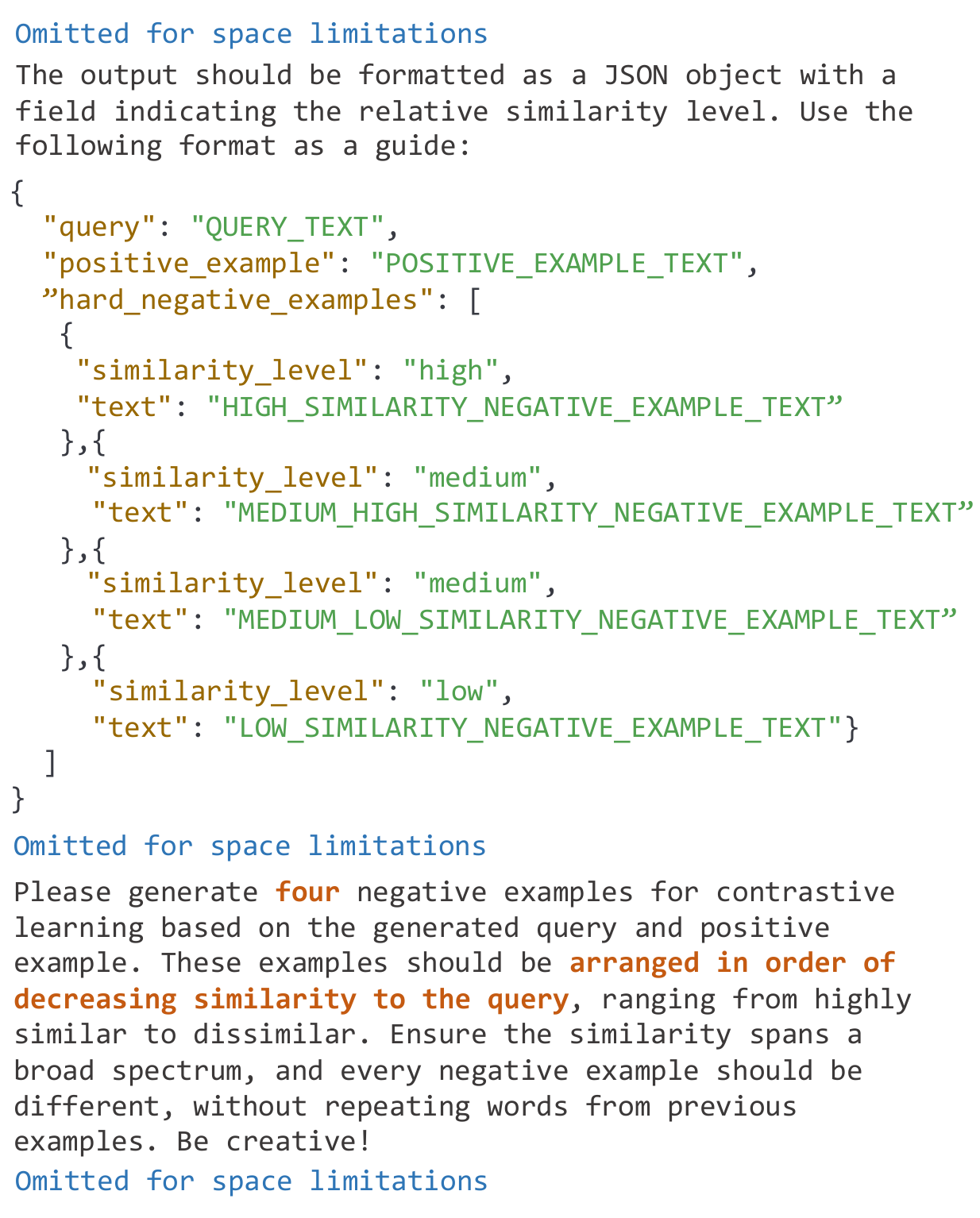}
  \caption{The core template used for prompting LLMs to generate multi-granularity hard negatives. Due to space constraints, the full prompts are presented in Appendix \ref{sec:appendix_full_prompts}. }
  \label{fig:template}
\end{figure}

We use the template illustrated in Figure \ref{fig:template} to constrain the synthesizing format. Differing from \citet{wang2024improving}, our approach prompts LLM to simultaneously generate multiple hard negative samples $\{\mathcal{N}^S_k\}$ for a single query  $\mathcal{Q}^S
$. These negative samples are ranked based on their similarity to the query, arranged in descending order from highest to lowest. This approach allows large models to enforce similarity constraints when generating hard negative samples, effectively mitigating the uncontrolled variation in the hard negative sample similarity during the synthetis process across different query samples. The effectiveness of this approach is demonstrated in Section \ref{sec:further_analysis} similarity statistics and a detailed case study.

In addition to synthetic data, we also incorporate public retrieval datasets when training the text embedding model, which include $\{(\mathcal{Q}^R, \mathcal{P}^R)\}$ sample pairs. To integrate coarse-to-fine hard negative samples into the subsequent training process, we regenerate the negative samples of the retrieval dataset, denoted as $\{\mathcal{N}^R_k\}$, by querying LLM using the same multi-granularity approach. The refined retrieval dataset, $\mathcal{T}^R = \{(\mathcal{Q}^R, \mathcal{P}^R, \{\mathcal{N}^R_k\})\}$, is then combined with the synthetic dataset $\mathcal{T}^S$ to form the complete training dataset $\mathcal{T}$.

With the multi-granularity dataset $\mathcal{S}$, we adopt a curriculum learning strategy for supervised training of the text embedding model. By adjusting the difficulty level $k$ of the hard negatives, the model progressively learns from coarse-grained to fine-grained distinctions, achieving more stable and effective embeddings. Specifically, during the supervised learning process, we gradually feed the negative samples into the model from $\{\mathcal{N}_4\}$ to $\{\mathcal{N}_1\}$, allocating equal proportions (25\% each) to four difficulty levels. Further analysis in Section \ref{sec:ablation-curriculum} demonstrates the effectiveness of the proposed scheduling method.

\subsection{Anchor Token Aware Pooling}
After obtaining the hidden states of multiple tokens from the model's final layer, an appropriate approach is required to derive the sentence representation vector $v$. In the widely adopted mean pooling method, the last hidden states of all tokens are averaged to form a sentence vector representation. This approach can result in the dilution of key information in the text, as non-critical tokens are averaged along with the more significant ones. 

The proposed ATA pooling method aims to assign greater weight to anchor tokens \citep{wang2023label}, which aggregate more semantic information compared to other tokens. This approach allows the model to adaptively allocate weights to the parts that are most pertinent to the task, resulting in a more effective pooling operation.

Motivated by \citet{huang2024opera}, we calculate the attention weight along the \texttt{Key} dimension of the attention matrix to identify anchor tokens with stronger representational capabilities. Compared to traditional mean pooling, this allows us to assign higher weights to anchor tokens and more effectively filter out trivial tokens that do not contribute to the semantic information.

Specifically, let $\mathbf{A}_\mathcal{L}^h$ denote the attention matrix for the attention head $h$ in the model’s final layer, and let $a^h_{ij}$ represent the corresponding value of $\mathbf{A}_\mathcal{L}^h[i][j]$, which indicates the attention score between \texttt{Query} $i$ and \texttt{Key} $j$. We define the anchor weight of each \texttt{Query} as follows:

\begin{equation}
  \label{eq:ata_pooling_1}
    \mathbf{w}_i = \sum_{h=1}^H \sum_{j=1}^K \log(a^h_{ij} \cdot K + 1),\ i\in[1,\dots,K]
\end{equation}

where $K$ represents the length of the token sequence, and $H$ denotes the number of attention heads. We multiply $a^h_{ij}$ by $K$ because, while the sum of attention weights in the query dimension remains constant (i.e. $\sum_{i=1}^K a^h_{ij} = 1$), the expected value of each individual element decreases to $\frac{1}{K}$ as the sentence length increases. Multiplying by the sentence length helps to ensure stability across different sentence lengths by maintaining consistent scaling in subsequent computations.

After obtaining the anchor weights, we normalize them by applying a linear weight adjustment along the \texttt{Query} dimension to compute the weight corresponding to each token, denoted as $\mathbf{\tilde{w}}_i$, such that the sum of all $\mathbf{\tilde{w}}_i$ equals 1:

\begin{equation}
  \label{eq:ata_pooling_2}
    \mathbf{\tilde w}_i = \frac{\mathbf{w}_i}{\sum_{j=1}^K \mathbf{w}_j},\ i\in[1,\dots,K]
\end{equation}

We then apply the normalized weights $\mathbf{\tilde{w}}_i$ to reweight the hidden states $\mathbf{H}_\mathcal{D}\in \mathbb{R}^{K\times\mathrm{hid\_dim}}$ and obtain the final sentence embedding $\mathbf{v}$:

\begin{equation}
  \label{eq:ata_pooling_3}
    \mathbf{v} = \sum_{i=1}^{K}\mathbf{\tilde{w}}_i\times\mathbf{H}_\mathcal{D}[i]
\end{equation}

\section{Experiments}

\subsection{Data Synthetic Details}

To ensure a fair comparison with \citet{wang2024improving}, we generated an equivalent volume of synthetic data, maintaining a consistent total token consumption of 180M.  In order to minimize the costs associated with data generation, we utilized the APIs of GPT-4o and DeepSeek-V2. We observed that compared to GPT-4o, DeepSeek-V2 is relatively less creative when generating the potential tasks in stage 1 and tends to produce repetitive negative samples during stage 2. Consequently, we relied on GPT-4o to complete all stage 1 generation processes. In stage 2, we initially used GPT-4o to generate a sufficient amount of data, which was then input as seeds into DeepSeek-V2. Leveraging the data cache provided by the DeepSeek-V2 API, introducing additional seeds as input did not incur significant additional costs. The distribution of synthetic data across different task types is detailed in Appendix \ref{sec:appendix_data_statistics}.

The retrieval dataset used in supervised learning was curated by \citet{springer2024repetition}, which consist of approximately 1.5 million samples, covering a variety of languages and retrieval scenarios. In line with LLM2Vec \citep{behnamghader2024llm2vec}, we only used about one-third of the curated retrieval dataset. For more details on the dataset composition, please refer to Appendix \ref{sec:appendix_public_training_data}.

\begin{table*} [ht]
  \centering
  \resizebox{\textwidth}{!}{
    \setlength{\tabcolsep}{4pt}
      \begin{tabular}{cccccccccc}
        \hline
         & \textbf{FT. Data} & \textbf{Class.(12)} & \textbf{Clust.(11)} & \textbf{Pair.(3)} & \textbf{Rerank.(4)} & \textbf{Retr.(15)} & \textbf{STS(10)} & \textbf{Summ.(1)} & \textbf{Avg.(56)} \\
        & Sample Num & Acc. & V-Meas. & AP & MAP & nDCG@10 & Spear. & Spear.
        \\
        \hline
        \multicolumn{10}{c}{\textbf{Models trained with synthetic data only}} \\
        \hline
        Mistral$_{\text{gpt-4o}}$ \citep{chen2024littlegiantssynthesizinghighquality} & 230K & 77.7 & 47.7 & 83.9 & 58.7 & 46.7 & 80.9 & 30.7 & 62.2 \\
        Gecko$_{\text{o1b-768}}$ \citep{lee2024gecko}& 6.6M & 70.3 & 46.8 & \underline{86.2} & 57.6 & \textbf{53.2} & \textbf{83.1} & \textbf{32.2} & 62.6 \\
        E5$_{\text{mistral-7b}}$ \citep{wang2024improving} & 500K & 78.2 & \textbf{50.5} & 86.0 & 59.0 & 46.9 & 81.2 & \underline{31.9} & 63.1 \\
        SPEED \citep{chen2024littlegiantssynthesizinghighquality} & 920K & \underline{78.3} & 48.6 & \textbf{86.3} & \underline{59.8} & 48.1 & \underline{82.6} & 31.7 & \underline{63.4} \\
        MGH(Ours) & 310K & \textbf{78.6} & \underline{49.7} & 86.1 & \textbf{60.1} & \underline{51.2} & 82.3 & 31.6 & \textbf{64.5} \\
        \hline
        \multicolumn{10}{c}{\textbf{Models trained with synthetic data \& public available retrieval data}} \\
        \hline
        GTR$_{\text{xxl}}$ \citep{ni-etal-2022-large} & 662K & 67.4 & 42.4 & 86.1 & 56.7 & 48.5 & 78.4 & 30.6 & 59.0 \\
        text-embedding-3$_{\text{large}}$  \footnotemark  & - & 75.5 & 49.0 & 85.7 & 59.2 & 55.4 & 81.7 & 29.9 & 64.6 \\
        jina-embed \citep{sturua2024jinaembeddingsv3multilingualembeddingstask} & - & \textbf{82.6} & 45.3 & 84.0 & 58.1 & 53.9 & \textbf{85.8} & 29.7 & 65.5 \\
        Gecko$_{\text{o1b-768}}$ \citep{lee2024gecko} & >6.6M & \underline{81.2} & 47.5 & 87.6 & 58.9 & 55.7 & 85.1 & \textbf{32.6} & 66.3 \\
        E5$_{\text{mistral-7b}}$ \citep{wang2024improving}& 1.8M & 78.5 & \textbf{50.3} & \textbf{88.3} & \underline{60.2} & \underline{56.9} & 84.6 & \underline{31.4} & \underline{66.6} \\
        SPEED \citep{chen2024littlegiantssynthesizinghighquality}& 2.2M & 78.4 & 49.3 & \underline{88.2} & \textbf{60.8} & 56.5 & \underline{85.5} & 31.1 & 66.5 \\
        MGH(Ours) & 820K &  78.8 & \underline{50.1} & 87.9 & 59.8 & \textbf{57.5} & 85.6 & 31.3 & \textbf{67.0}\\
        \hline
      \end{tabular}
    }
    \caption{\label{tab:performance-comparison-1}
    Full MTEB benchmark performance comparison of different synthesis models, with training conducted on \textit{synthetic data only} and \textit{both synthetic and public retrieval dataset}. The highest performances are highlighted in bold, while the second-highest are indicated with underlines. Numbers in parentheses in column headers indicate the number of subtasks within each task category. \textit{FT. Data Sample Num.} refers to the number of sample pairs used for training. 
  }
\end{table*}

\subsection{Experimental Details}
\label{sec:experimental_details}

\textbf{Model Setup.} To validate the effectiveness of our proposed fine-grained data synthesis framework and the ATA pooling method, we perform experiments following the open-sourced LLM2Vec \citep{behnamghader2024llm2vec} model, while replacing the supervised training stage with our method. Specifically, we adopt Mistral-7B-Instruct-v0.2 \citep{jiang2023mistral} as the base model. We then transform the model's attention pattern from causal to bidirectional and integrate the LoRA weights trained on the masked next-token prediction task introduced in LLM2Vec, enabling the model to better adapt to bidirectional attention patterns. This setup serves as the starting point for our model.

\textbf{Supervised Training.} We adopt the standard InfoNCE loss, with both in-batch negatives and hard negatives utilized for training. To ensure a fair comparison, the prompt template follows \citet{wang2024improving}, as illustrated in Appendix \ref{sec:appendix_public_training_data}. Supervised training on the full dataset is conducted for 1600 steps, while training on public retrieval dataset is performed for 1000 steps, with a batch size of 64 and gradient accumulation of 8 in both cases. All training was performed on a single 80GB H100 GPU, taking approximately 32 hours to complete 1600 training steps. Further training hyperparameters are represented in Appendix \ref{sec:appendix_hyperparameters}. 

\textbf{Evaluation.} We assess our model on the widely used MTEB benchmark \citep{muennighoff2023mteb}, which encompasses a wide variety of text embedding tasks across different scenarios and domains. The benchmark includes 56 English embedding tasks organized into 7 categories: classification (12), clustering (11), pair classification (3), reranking (4), retrieval (15), semantic textual similarity (10), and summarization (1).

\subsection{Main Results}

Table \ref{tab:performance-comparison-1} presents a comparison of the performance of our proposed method against existing data synthesis approaches on the MTEB dataset. The results underscore the effectiveness of our data generation framework, demonstrating superior performance in both the synthetic-only and full-data settings. In particular, on the more challenging retrieval tasks, the full-data results achieved the best performance, underscoring the efficacy of our synthesis method.

\footnotetext{\href{https://platform.openai.com/docs/guides/embeddings}{https://platform.openai.com/docs/guides/embeddings}}

To evaluate the effectiveness of the proposed ATA pooling method, we compare our model with previous approaches that leverage LLMs for text embedding. However, previous research has indicated that incorporating the MTEB dataset's training split during the supervised training process introduces a significant amount of MTEB-related data, thereby increasing the risk of over-fitting \citep{li2024making}. Therefore, we opted not to include the second training phase proposed by NV-Embed, which utilizes the MTEB training split for a continuous training.

\begin{table}[ht]
\centering
\resizebox{\columnwidth}{!}{%
\begin{tabular}{lc}
\hline
\textbf{Model} & \textbf{MTEB Score} \\ \hline
SGPT \citep{muennighoff2022sgptgptsentenceembeddings} & 58.93 \\
UDEVER-bloom-7b \citep{zhang2023languagemodelsuniversalembedders} & 60.63 \\
ECHO \citep{springer2024repetition} & 64.68 \\
LLM2Vec \citep{behnamghader2024llm2vec} & 64.80 \\
NV-Embed \citep{lee2024nv}  & 64.18 \\ 
bge-large-en-v1.5 \citep{xiao2023bge} & 64.23 \\
bge-en-icl w/o icl \citep{li2024making} & 64.83 \\
bge-en-icl w/ icl \citep{li2024making}& 66.08 \\
MGH(Ours) w/o icl & 65.87 \\
MGH(Ours) w/ icl &  \textbf{66.43} \\
\hline
\end{tabular}
}
\caption{Performance comparison of MTEB scores for different models trained on publicly available retrieval corpora, without introducing MTEB training split during training.}
\label{tab:performance-comparison-2}
\end{table}

Table \ref{tab:performance-comparison-2} compares the performance of our approach with existing models trained exclusively on publicly available retrieval data. Our model achieves state-of-the-art performance under the MTEB training split free setting, demonstrating the effectiveness of the ATA pooling method.



\section{Ablation Study}

\subsection{Data Synthesis and Training Strategies}
\label{sec:ablation-curriculum}
In this section, we evaluate the effectiveness of the proposed MGH synthesis framework through an ablation study on a subset of the MTEB benchmark, as used in \citet{springer2024repetition}\footnote{Conducting a full evaluation on the MTEB dataset is computationally expensive, requiring over 200 hours on a single H100 GPU. Therefore, a subset of the dataset was selected for this study. Details of the subset composition can be found in Appendix \ref{sec:appendix_mteb_subset}.}. We evaluate the impact of data training order on supervised learning through the following experimental settings, analyzing how the results evolve throughout the training process over 1600 steps:

\begin{figure}[t]
  \centering
  \includegraphics[width=\columnwidth]{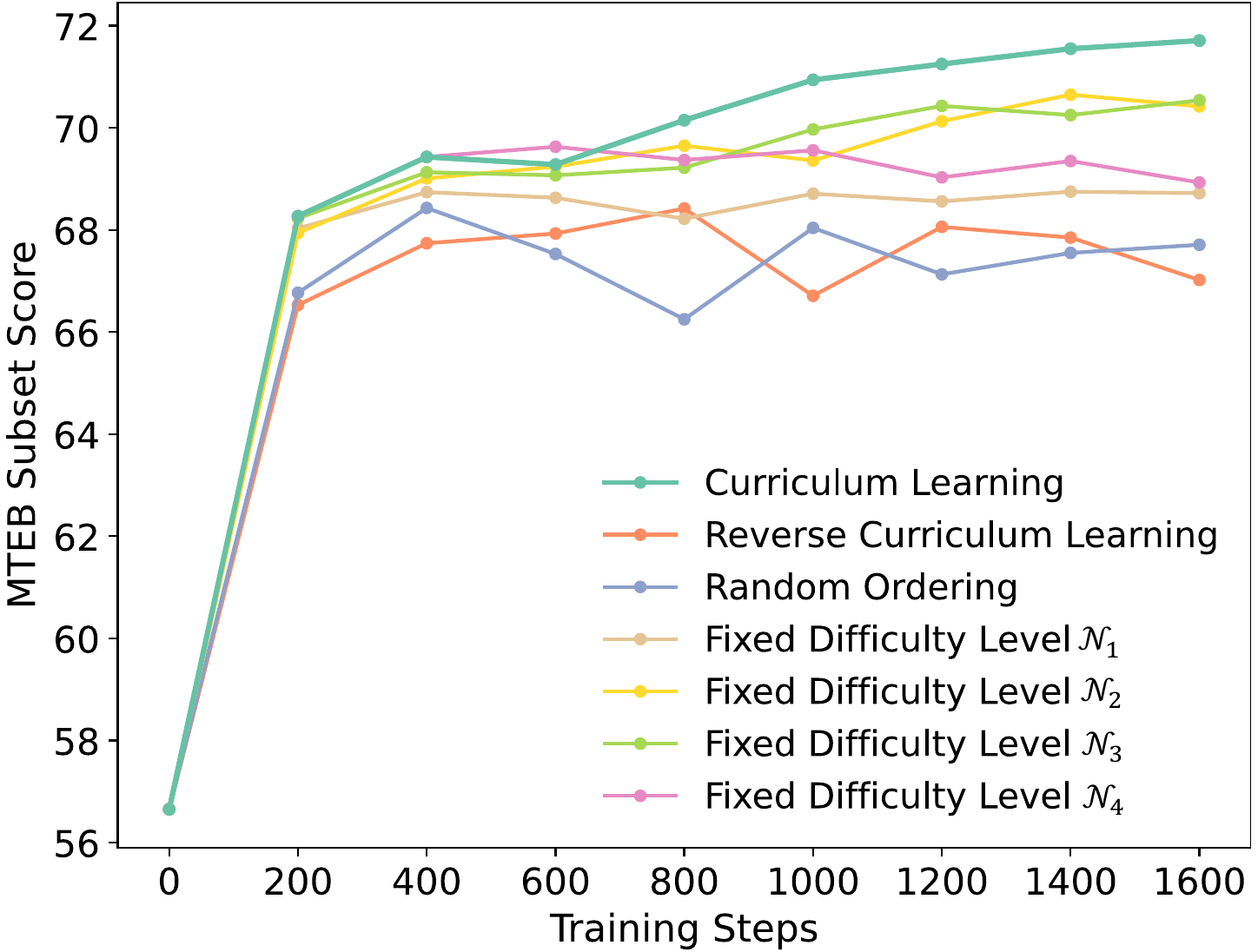}
  \caption{Trend of MTEB subset scores during supervised training across four experimental settings.}
  \label{fig:ablation}
\end{figure}

\begin{figure*}[t]
  \centering
  \includegraphics[width=\textwidth]{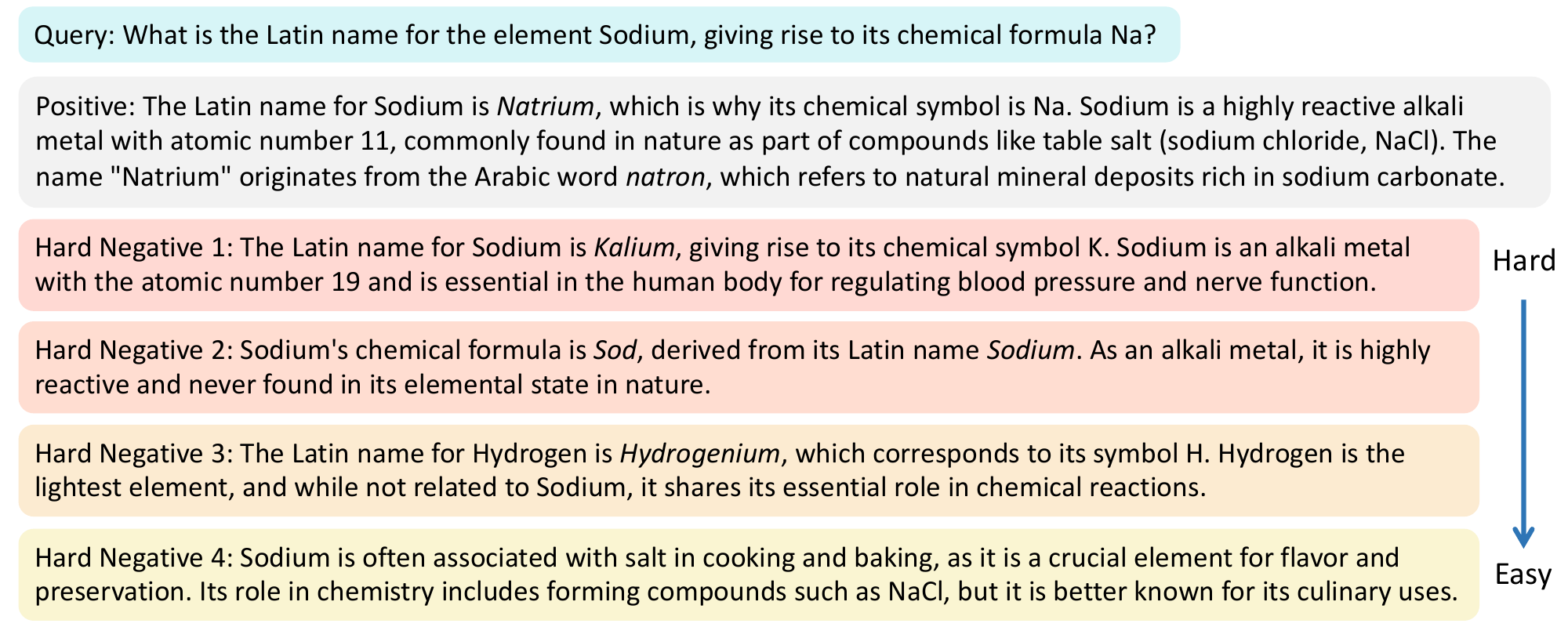}
  \caption{An example of a multi-granularity hard negative synthesis, randomly selected for illustration.}
  \label{fig:case_study}
\end{figure*}

\begin{enumerate}
    \setlength{\itemsep}{0pt}
    \setlength{\parskip}{0pt}
    \item \textbf{Curriculum Learning:} Progressing from easier to more challenging hard negatives, as adopted in our main approach.
    \item \textbf{Reverse Curriculum Learning:} Progressing hard negative samples from harder to easier.
    \item \textbf{Random Ordering:} Randomly inputting hard negative examples of varying difficulty.
    \item \textbf{Fixed Difficulty Level:} Consistently using hard negatives of a fixed difficulty level.
\end{enumerate}

The results in Figure \ref{fig:ablation} demonstrate that the curriculum learning strategy adopted by MGH not only achieves the best performance but also maintains greater stability during training. In contrast, the random ordering strategy predictably exhibits fluctuations in the results, while the reverse curriculum learning method also fails to yield satisfactory training outcomes, as the model struggles to adapt to the reversed difficulty progression.

Additionally, none of the four fixed difficulty settings outperformed the curriculum learning approach. Among them, maintaining a difficulty level of 2 or 3 yielded relatively better results, while using excessively low or high difficulty levels failed to converge to optimal outcomes. This suggests that overly simple negative examples may prevent the model from learning useful knowledge, and excessively difficult synthetic negatives may hinder the model’s ability to distinguish them from positive examples at early training stages.

\subsection{Pooling Methods}
\label{sec:pooling_methods}

This section presents a detailed comparison of four pooling methods employed in embedding tasks, namely mean pooling, last token pooling, NV-Embed pooling \citep{lee2024nv}, and the proposed ATA pooling. The ablation experiments leveraged publicly available retrieval datasets only, adhering to the hyperparameter settings outlined in Section \ref{sec:experimental_details}. The evaluation is conducted using the full MTEB benchmark, with Table \ref{tab:ablation-pooling} summarizing their respective performances.

\begin{table}[ht]
\centering
\resizebox{0.35\textwidth}{!}{
\begin{tabular}{lc}
\hline
\textbf{Pooling Method} & \textbf{MTEB Score} \\ \hline
mean pooling & 65.41 \\
last pooling & 64.97 \\
NV-Embed pooling & 65.80 \\
ATA pooling (Ours)  & \textbf{65.87} \\ 
\hline
\end{tabular}
}
\caption{Performance comparison of MTEB scores for different pooling methods trained on publicly available retrieval corpora.}
\label{tab:ablation-pooling}

\end{table}

The results suggest that conventional mean pooling and last token pooling yield subpar performance in text embedding tasks using bidirectional models. On the other hand, both the NV-Embed pooling and ATA pooling methods demonstrate favorable results, validating the necessity of adaptive reweighting for the last hidden states.

\section{Further Analysis}
\label{sec:further_analysis}

\subsection{Difficulty of Synthesized Hard Negatives}

To assess the granularity of generated negative samples, we use the model trained exclusively on publicly available retrieval corpora, as described in Section \ref{sec:pooling_methods}, to evaluate the similarity between the four levels of synthesized $\mathcal{N}_k$ and the query $\mathcal{Q}$. Since this model was not exposed to the synthesized negative samples during training, it serves as an unbiased evaluator to assess the effectiveness of our MGH framework in generating negative samples with different difficulty levels.

\begin{table}[ht]
\centering
\resizebox{0.45\textwidth}{!}{
\begin{tabular}{cc}
\hline
\textbf{Negative Granularity} & \textbf{Cosine Similarity} \\ \hline
$\mathcal{N}_1$ (Hardest) & 0.881 \\
$\mathcal{N}_2$ (Medium) & 0.857 \\
$\mathcal{N}_3$ (Medium) & 0.845 \\
$\mathcal{N}_4$ (Easiest) & 0.793 \\ 
\hline
\end{tabular}
}
\caption{Average cosine similarity between queries and negative samples at different difficulty levels.}
\label{tab:negative_similarity}

\end{table}

The experimental results in Table \ref{tab:negative_similarity} demonstrate that average cosine similarity between the negative samples and the query increases progressively from $\mathcal{N}_4$ to $\mathcal{N}_1$. This aligns with our initial hypothesis that $\mathcal{N}_1$ (hardest) negative samples should be more similar to the query than $\mathcal{N}_4$ (easiest), indicating that our MGH framework effectively generates increasingly challenging negative samples at higher difficulty levels. 

\subsection{Case Study}

\paragraph{\textbf{How does MGH enhance hard negative sample quality?}} We illustrate this through the example presented in Figure \ref{fig:case_study}, which demonstrates how MGH effectively improves the quality of hard negative samples by leveraging multi-granularity similarity constraints.

In this example, the data synthetic model is tasked with generate negative samples for the Latin name of the element \textit{Sodium}. The GPT-4o model used in this case selects \textit{Potassium}, which shares similar chemical properties with \textit{Sodium}, as the most challenging hard negative example. Subsequently, the model generates a fake Latin name for \textit{Sodium} as a moderately confusable negative sample, followed by answering element \textit{Hydrogen} as a more distinguishable example. While the first three negative samples involve answering the Latin names of chemical elements, the last simplest negative sample generated by the model focuses on \textit{Sodium} but lacks any reference to its Latin name. 

As demonstrated in the example above, the MGH approach effectively distills world knowledge from LLMs, enabling the generation of multiple negative samples with varying granularities. As the examples progress from challenging to simple, the synthetic model's outputs range from showing subtle differences in detail to being more easily distinguishable. In this process, the subsequent negative samples are adjusted based on the previously generated prefix, enabling a dynamic progression of negative sample difficulty, further enhancing the quality of negative sample generation. 

\begin{figure}[t]
  \centering
  \includegraphics[width=0.9\columnwidth]{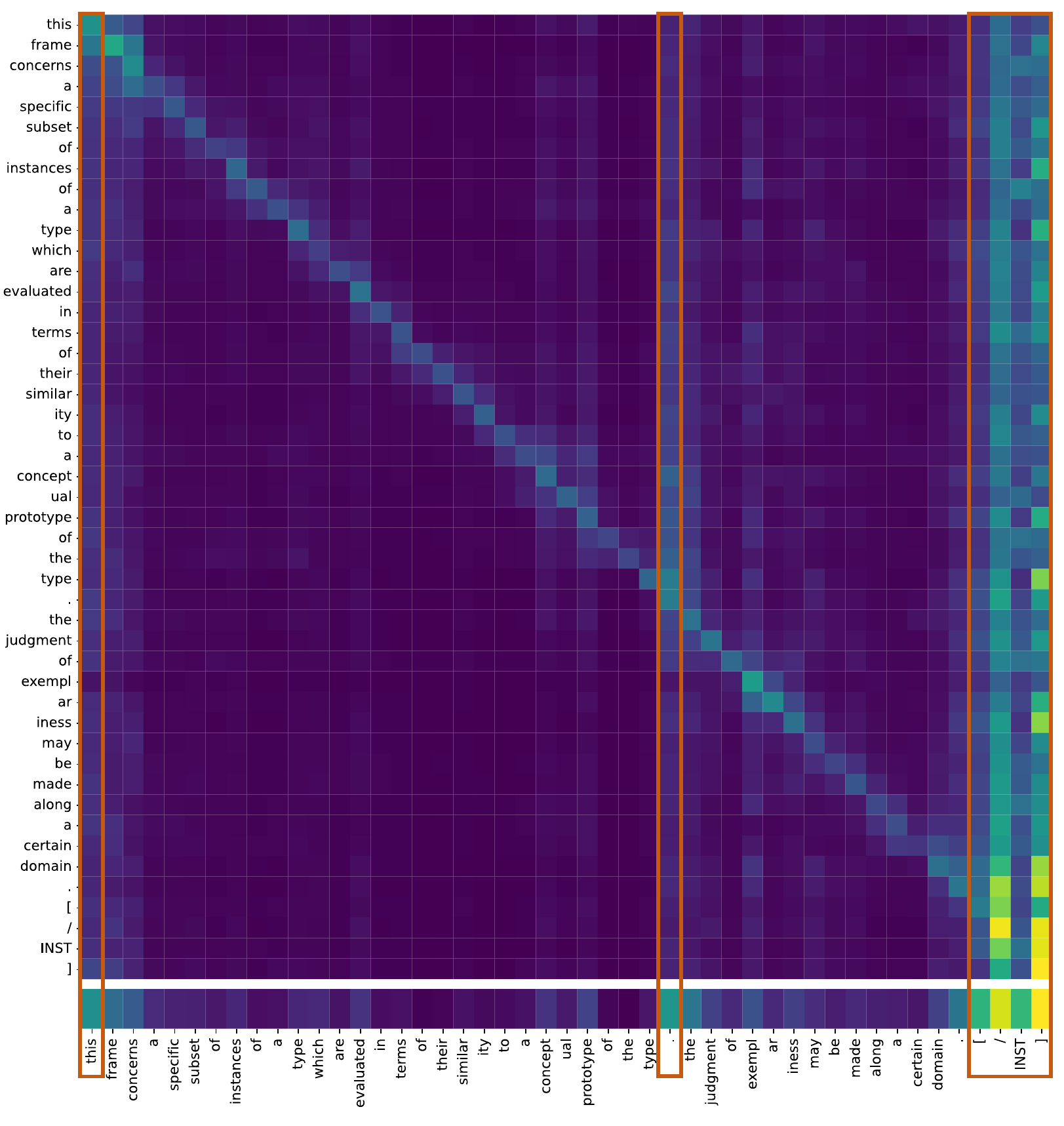}
  \caption[]{The upper part illustrates the summed results of the model's final layer attention weights across 32 attention heads\footnotemark, while the lower part shows the corresponding ATA weights for each token. Example is randomly selected in STS13 evaluation split.}
  \label{fig:case_study_ata}
\end{figure}

\paragraph{\textbf{How does ATA reweight using aggregation pattern?}} As shown in the example from Figure \ref{fig:case_study_ata}, the aggregation pattern still remains when the base model is transformed from causal to bidirectional attention. The figure illustrates three prominent anchor tokens: the initial token, the punctuation between the two sentences, and the \texttt{[INST]} template appended to the end of the sentence. Accordingly, these anchor tokens receive higher weight values in the ATA weight calculation, contributing more significantly to the subsequent computation of text embeddings.

Through observations of numerous examples, we found that most samples allocate a greater proportion of the ATA weight to the three anchor patterns mentioned above, with particular emphasis on the \texttt{[INST]} token at the sentence's end. Therefore, the ATA pooling method captures the important last token while also dynamically identifying key positions within the preceding text. This approach not only mitigates the stability issues associated with relying solely on the last token but also assigns greater weight to tokens that are essential for capturing the entire semantic meaning of the input sequence, thereby facilitating more effective embedding learning.

\subsection{Cost of Synthetic Data}

Although our model entails additional tokens to generate multiple hard negatives per synthetic sample, the cost is offset by maintaining the same token consumption (180M) as \citet{wang2024improving}, which results in fewer synthetic samples being generated.  Under this fair comparison, the model trained with our MGH strategy outperforms previous state-of-the-art results, demonstrating that our method is more effective under the same token consumption.


\section{Related Work}

\paragraph{\textbf{LLM Based Text Embedding Models}} In recent years, as decoder-only models have scaled up in terms of parameters and training data, researchers have explored the possibility of transforming next-token prediction models into effective text embedding models through continued training. \citet{neelakantan2022textcodeembeddingscontrastive} was the first to apply the GPT-3 model to text embedding tasks, leveraging the \texttt{<EOS>} token as the representation vector. Subsequent work by \citet{ma2024fine} employed a similar last-token pooling method, fine-tuning the LLaMA-2 model. 

\footnotetext{For space limitations, the individual attention maps from all 32 attention heads are provided in Appendix \ref{sec:appendix_attention_matrices}.}

However, the autoregressive training objective imposes an inherent limitation on the model's performance, as the causal attention mask prevents earlier tokens from accessing subsequent tokens. SGPT \citep{muennighoff2022sgptgptsentenceembeddings} addressed this limitation by linearly assigning more weight to tokens at later positions, a strategy subsequently adopted by E5$_{\text{mistral-7b}}$ \citep{wang2024improving}. LLM2Vec \citep{behnamghader2024llm2vec} transformed the model from causal to bidirectional attention by employing a masked next-token prediction approach, followed by mean pooling for supervised learning. Recent work by NV-Embed \citep{lee2024nv} introduced an additional cross-attention layer for hidden state pooling, simultaneously removing the causal mask. Additionally, Echo embeddings \citep{springer2024repetition} repeated a text twice and used the second instance to compute the representation vector. In this work, we attempt to address the insufficient pooling problem in LLM based embedding models by an adaptive weighting strategy using the model's aggregation pattern.

\paragraph{\textbf{Data Synthesis for Embedding Models}} High-quality data is crucial for training effective text embedding models. Previous studies \citep{nogueira2019document, wang-etal-2023-query2doc} have explored expansion-based approaches to augment document and query data. With the advancements in large language model (LLM) capabilities, recent research has focused on leveraging LLMs to generate large amounts of high-quality supervised training data \citep{wang2024improving, jeronymo2023inparsv2largelanguagemodels, sturua2024jinaembeddingsv3multilingualembeddingstask}. In domain-specific retrieval, studies \citep{dai2022promptagatorfewshotdenseretrieval, khramtsova2024leveraging} have shown that LLM-generated query-document pairs significantly improve embedding quality in domain-specific retrieval tasks. Additionally, Gecko \citep{lee2024gecko} curates web data to enable LLMs to produce high-quality synthetic samples. Our work focuses on how to enable large models to generate multi-granularity synthetic negative examples, and achieves more efficient and stable text embedding model training by controlling the training difficulty.

\section{Conclusion}

In this work, we evaluate the importance of hard negative granularity when training text embedding models using contrastive learning. Through the proposed MGH synthesis framework, we generate diverse negative samples at varying levels of similarity, enabling the embedding model to learn more nuanced semantic representations by coarse-to-fine curriculum learning approach. Experimental results demonstrate that our methods achieve state-of-the-art performance on the MTEB benchmark, outperforming existing synthesis strategies both with synthetic-only and combined datasets. Additionally, our proposed ATA pooling method effectively leverages the aggregation patterns inherent in large language models, improving sentence pooling efficacy without introducing extra parameters. Ablation studies confirm the effectiveness of our MGH framework and ATA pooling method in enhancing text embedding model performance and training stability.

\section*{Limitations}

Despite the effectiveness of our method, there are several limitations that should be acknowledged: (1) Due to the costs associated with API usage, we limited the synthetic data generation to the same token volume as used in previous studies \citep{wang2024improving}. This constraint prevented us from exploring whether a larger synthetic dataset could further improve the performance of the text embedding model.  We leave this exploration to future work, where more extensive synthetic data could be generated to assess the scalability and potential performance gains. (2) To facilitate comparisons with prior work \citep{behnamghader2024llm2vec, springer2024repetition}, we used Mistral-7b-v0.2-Instruct as our base text embedding model. Given the continuous advancements in 7b-level models, we plan to investigate more powerful models as our base embedding model in our future work.

\section*{Acknowledgement}
This work was supported in part by National Key Research and Development Program of China under Grant No. 2020YFA0804503, National Natural Science Foundation of China under Grant No. 62272264, ByteDance Doubao Large Model Fund Project under Grant No. CT20240909109354, and National Natural Science Foundation of China under Grant No. 62402294.

\bibliography{custom}

\appendix

\section{Experimental Details for Data Synthesis}
\subsection{Prompts for Data Synthesis}
\label{sec:appendix_full_prompts}

To enable large models to generate multiple hard negatives with varying granularities, we extend and refine the prompt template proposed by \citet{wang2024improving}.  Specifically, we modify the original template to guide the model in producing negative samples with different levels of similarity to the query, thereby enhancing the diversity and difficulty of the generated data. Table \ref{tab:app_short_long} illustrates the complete prompt template used to generate short-long match tasks as an example.

\subsection{Details of Task Categories}
\label{sec:appendix_details_of_categories}

During the task brainstorming process in synthesis stage 1, we followed \citet{wang2024improving} to systematically classify the potential tasks into two distinct categories: asymmetric tasks and symmetric tasks. The fundamental distinction lies in the semantic relationship between queries and their corresponding positive documents. In asymmetric tasks, the query and its relevant document exhibit semantic relevance but are not paraphrases of one another. Conversely, symmetric tasks are characterized by query-document pairs that preserve semantic equivalence through different linguistic formulations.

\subsection{Statistics on Synthesized Data}
\label{sec:appendix_data_statistics}

\begin{figure}[h]
  \centering
  \includegraphics[width=0.90\columnwidth]{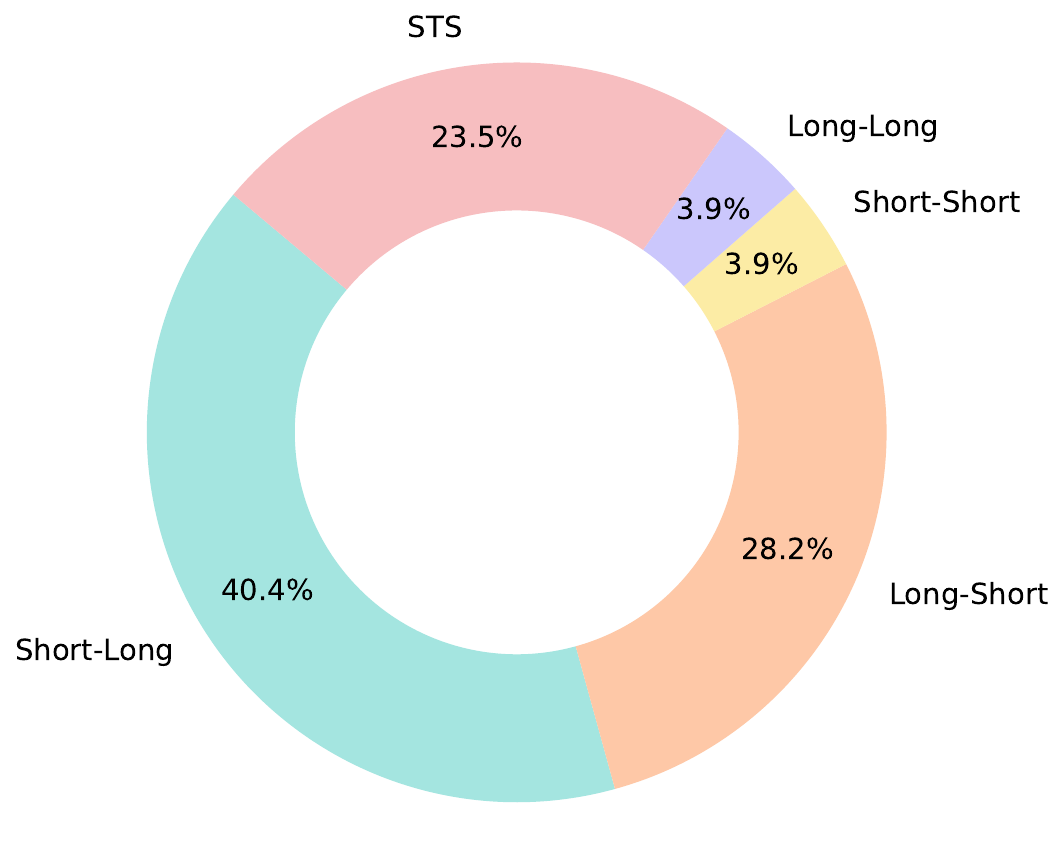}
  \caption{Distribution of the task categories in the synthetic data}
  \label{fig:pie_chart}
\end{figure}

Figure \ref{fig:pie_chart} illustrates the distribution of synthetic data across five different task types. In line with \citet{wang2024improving}, we adopted the same task ratio allocation, where Short-Long, Long-Short, and STS comprise the majority of the data, while Long-Long and Short-Short account for relatively smaller proportions.

\section{Experimental Details for Supervised Training}
\label{sec:appendix}

\subsection{Hyperparameters}
\label{sec:appendix_hyperparameters}

We present the hyperparameters involved in the supervised training in Table \ref{tab:hyperparameters}. The \textit{max sequence length} specifies that any text sequence exceeding this number of tokens is truncated in our text embedding model.

\begin{table}[t]
\centering
\resizebox{0.35\textwidth}{!}{
\begin{tabular}{lc}
\hline
\textbf{Hyperparameter} & \textbf{Value} \\ \hline
Batch Size & 64 \\
Gradient Accumulation Steps & 8 \\
Learning Rate & 2e-5 \\
Max Sequence Length & 512 \\
LoRA rank & 16 \\
LoRA $\alpha$ & 32 \\ 
Optimizer & Adam \\ \hline
Training Steps - Synthetic Only & 600 \\
Training Steps - Public Only & 1000 \\
Training Steps - Synthetic \& Public & 1600 \\ \hline
Warmup Steps - Synthetic Only & 200 \\
Warmup Steps - Public Only & 300 \\
Warmup Steps - Synthetic \& Public & 300 \\

\hline
\end{tabular}
}
\caption{Hyperparameters used in the experiments}
\label{tab:hyperparameters}
\end{table}

\subsection{Training Time}
When training for 1600 steps on a single H100 GPU, the time required with and without ATA pooling was 32.45 hours and 32.40 hours, respectively, demonstrating that out proposed ATA pooling strategy doesn't adds additional computational overhead compared to the LLM backbone. The slight variations in training and inference time are likely due to normal fluctuations in GPU workload and system scheduling.

\subsection{Public Retrieval Datasets}
\label{sec:appendix_public_training_data}

We follow the training data setup from previous work \citep{springer2024repetition,behnamghader2024llm2vec}, adopting the dataset configuration used by \citet{wang2024improving}, which includes the following datasets: ELI5 \citep{fan2019eli5} (sample ratio 0.1) , HotpotQA \citep{yang2018hotpotqa}, FEVER \citep{thorne2018fever}, MIRACL \citep{zhang2023miracl}, MS-MARCO \citep{bajaj2016ms} passage ranking (sample ratio 0.5) and document ranking (sample ratio 0.2), NQ \citep{karpukhin2020dense}, NLI \citep{gao2021simcse}, SQuAD \citep{karpukhin2020dense}, TriviaQA \citep{karpukhin2020dense}, Quora Duplicate Questions \citep{quora-question-pairs} (sample ratio 0.1), Mr-TyDi \citep{zhang2021mr}, DuReader \citep{qiu2022dureader}, and T2Ranking \citep{xie2023t2ranking} (sample ratio 0.5). The full supervised training data has approximately 1.5M training examples. The instructions applied to each dataset are in line with \citet{behnamghader2024llm2vec}, which are listed in Table \ref{tab:e5_instructions}.

\subsection{Similarity Function}

The cosine similarity was adopted as the similarity metric in Equation \ref{eq:info_nce}, which can be mathematically expressed as follows:

$$
\phi = \frac{{\bar u}\cdot{\bar v}}{||\bar u||\times||\bar v||}
$$

\input{Tables/prompt_short_long}

\input{Tables/e5_instructions}

\section{Experimental Details for Evaluation}
\subsection{Prompts for MTEB Evaluation}

For a fair comparison with previous work \citep{wang2024improving, behnamghader2024llm2vec, lee2024nv} evaluated on MTEB, we adopted the same set of prompt instructions used in their evaluations when assessing our model's performance. The instructions applied to each evaluation dataset are listed in Table \ref{tab:mteb-instructions}.

\input{Tables/mteb_instructions}

\subsection{Subset Used for Ablation Study}
\label{sec:appendix_mteb_subset}
To speed up evaluation in the ablation study, we follow \citet{springer2024repetition} by selecting a representative subset of the MTEB evaluation benchmark, which includes the following datasets: FiQA2018, SCIDOCS, SciFact, NFCorpus, TwitterSemEval2015, TwitterURLCorpus, ImdbClassification, AmazonReviewsClassification, TweetSentimentExtractionClassification, MTOPDomainClassification, TwentyNewsgroupsClustering, BiorxivClusteringS2S, MedrxivClusteringS2S, StackOverflowDupQuestions, AskUbuntuDupQuestions, SciDocsRR, BIOSSES, STS12, STS13, STS14, STS15, STS16, STS17, STS22, STSBenchmark, and SICK-R. 

\section{Additional Results}

\subsection{Full MTEB Results}
\label{sec:appendix_mteb_results}

In this section, we present the complete results for all 56 MTEB datasets across the three experimental settings of our main experiment: public retrieval data only, synthetic data only, and full data. The corresponding results are shown in Table \ref{tab:full-mteb-result}.

\input{Tables/full_mteb_result}

\subsection{Full Attention Matrices}
\label{sec:appendix_attention_matrices}

\begin{figure*}[t]
  \centering
  \includegraphics[height=0.90\textheight]{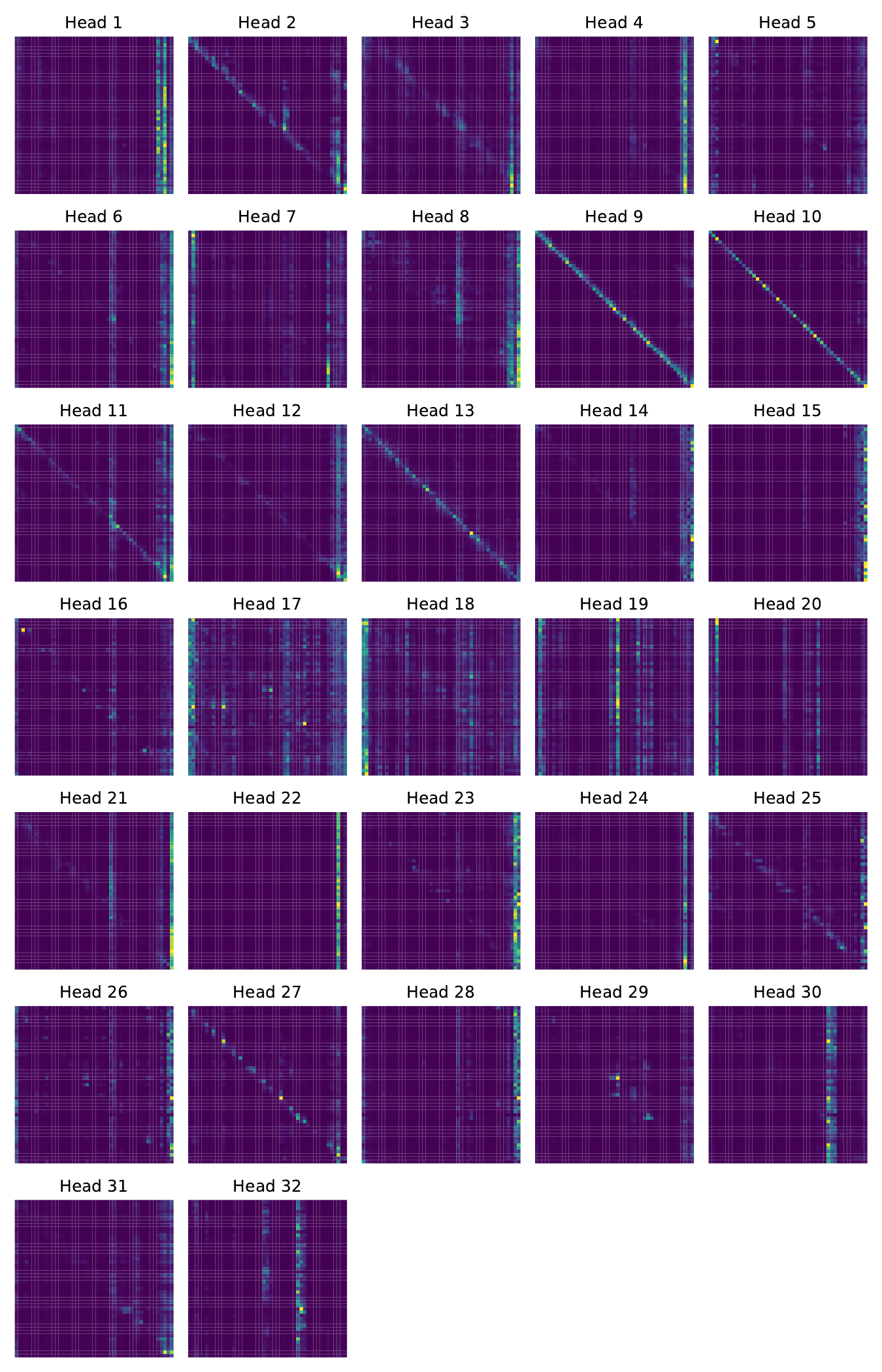}
  \caption{The individual attention matrices from all 32 attention heads in the last layer of the bidirectional model, obtained from a randomly selected example in the STS13 dataset.}
  \label{fig:case_study_ata_separate}
\end{figure*}

As shown in Figure \ref{fig:case_study_ata_separate}, after transforming the attention mask of the base model (i.e. Mistral-7B-Instruct-v0.2) from causal to bidirectional, the attention heads continue to exhibit distinct patterns, with some heads focus on tokens in their original positions, while others show higher attention scores across all query dimensions at the current position (e.g. Head 1, Head 6, Head 21 and Head 22). The latter pattern reflects the characteristics of anchor tokens, allowing for more effective aggregation of information from the entire sentence. Consequently, in our ATA pooling method, these attention heads are assigned with greater pooling weight.

\end{document}

%% file: Tables/prompt_short_long.tex
\begin{table*}[ht]
\centering
\small{\begin{tabular}{l}
\hline
\begin{tabular}[c]{@{}p{0.97\linewidth}@{}}Brainstorm a list of potentially useful text retrieval tasks.\\ \\ Here are a few examples for your reference:\\ - Retrieve relevant documents for a short keyword web search query that asks for weather information.\\ - Search for documents that answers a FAQ-style query on children's nutrition.\\ \\ Please adhere to the following guidelines:\\ - Specify what the query is, and what the desired documents are.\\ - Each retrieval task should cover a wide range of queries, and should not be too specific.\\ \\ Your output must always be a python list of strings only, with about 20 elements, and each element corresponds to a distinct retrieval task in one sentence. Do not explain yourself or output anything else. Be creative!\end{tabular} \\ \hline
\begin{tabular}[c]{@{}p{0.97\linewidth}@{}}You have been assigned a retrieval task: \{task\}\\ \\ Your mission is to write one text retrieval example for this task in the following JSON format. The JSON object must contain the following keys: \\ - "user\_query": a string, a random user search query specified by the retrieval task.\\ - "positive\_document": a string, a relevant document for the user query.\\ - "hard\_negative\_document": a list of strings, hard negative documents that only appears relevant to the query.\\

\\ The output should be formatted as a JSON object with a field indicating the relative similarity level of hard negative examples. Use the following format as a guide: \\ \\

\{
\\
\quad "user\_query": "QUERY\_TEXT",
\\
\quad "positive\_document": "POSITIVE\_EXAMPLE\_TEXT",
\\
\quad "hard\_negative\_document": [
\\
\quad\quad \{
\\
\quad\quad\quad "similarity\_level": "high",
\\
\quad\quad\quad "text": "HIGH\_SIMILARITY\_NEGATIVE\_EXAMPLE\_TEXT"
\\
\quad\quad \},\{
\\
\quad\quad\quad "similarity\_level": "medium",
\\
\quad\quad\quad "text": "MEDIUM\_HIGH\_SIMILARITY\_NEGATIVE\_EXAMPLE\_TEXT"
\\
\quad\quad \},\{
\\
\quad\quad "similarity\_level": "medium",
\\
\quad\quad "text": "MEDIUM\_LOW\_SIMILARITY\_NEGATIVE\_EXAMPLE\_TEXT"
\\
\quad\quad \},\{
\\
\quad\quad\quad "similarity\_level": "low",
\\
\quad\quad\quad "text": "LOW\_SIMILARITY\_NEGATIVE\_EXAMPLE\_TEXT"
\\
\quad\quad \}
\\
\quad ]
\\
\}

\\ Please adhere to the following guidelines:\\ - The "user\_query" should be \{query\_type\}, \{query\_length\}, \{clarity\}, and diverse in topic.\\ - All documents must be created independent of the query. Avoid copying the query verbatim. It’s acceptable if some parts of the "positive\_document" are not topically related to the query.\\ - All documents should be at least \{num\_words\} words long.\\ - The "hard\_negative\_document"  contains some useful information, but it should be less useful or comprehensive compared to the "positive\_document". Please generate four hard negative documents for contrastive learning based on the generated query and positive example. These examples should be arranged in order of decreasing similarity to the query, ranging from highly similar to dissimilar. Ensure the similarity spans a broad spectrum, and every negative example should be different, without repeating words from previous examples.\\ - Both the query and documents should be in \{language\}.\\ - Do not provide any explanation in any document on why it is relevant or not relevant to the query.\\ - Both the query and documents require \{difficulty\} level education to understand.\\ \\ Your output must always be a JSON object only, do not explain yourself or output anything else. Be creative!\end{tabular} \\ \hline
\end{tabular}}
\caption{Prompt template for the short-long matching task.
For placeholders,
``\emph{\{query\_type\}}'' $\in$ \{extremely long-tail, long-tail, common\},
``\emph{\{query\_length\}}'' $\in$ \{less than 5 words, 5 to 15 words, at least 10 words\},
``\emph{\{difficulty\}}'' $\in$ \{high school, college, PhD\},
``\emph{\{clarity\}}'' $\in$ \{clear, understandable with some effort, ambiguous\},
``\emph{\{num\_words\}}'' $\in$ \{50, 100, 200, 300, 400, 500\}.}
\label{tab:app_short_long}
\end{table*}

%% file: Tables/e5_instructions.tex
\begin{table*}[ht]
\centering
\resizebox{2.1\columnwidth}{!}{%
\begin{tabular}{l|p{0.8\linewidth}}
\hline
\textbf{Task Name} & \textbf{Instruction} \\ \hline
NLI & Given a premise, retrieve a hypothesis that is entailed by the premise \\
    & Retrieve semantically similar text \\
DuReader & Given a Chinese search query, retrieve web passages that answer the question \\
ELI5 & Provided a user question, retrieve the highest voted answers on Reddit ELI5 forum \\
FEVER & Given a claim, retrieve documents that support or refute the claim \\
HotpotQA & Given a multi-hop question, retrieve documents that can help answer the question \\
MIRACL & Given a question, retrieve Wikipedia passages that answer the question \\
MrTyDi & Given a question, retrieve Wikipedia passages that answer the question \\
MSMARCO Passage & Given a web search query, retrieve relevant passages that answer the query \\
MSMARCO Document & Given a web search query, retrieve relevant documents that answer the query \\
NQ & Given a question, retrieve Wikipedia passages that answer the question \\
QuoraDuplicates & Given a question, retrieve questions that are semantically equivalent to the given question \\
 & Find questions that have the same meaning as the input question \\
SQuAD & Retrieve Wikipedia passages that answer the question \\
T2Ranking & Given a Chinese search query, retrieve web passages that answer the question \\
TriviaQA & Retrieve Wikipedia passages that answer the question \\
\hline
\end{tabular}
}
\caption{The prompt instructions used for public retrieval datasets, following \citet{behnamghader2024llm2vec}}
\label{tab:e5_instructions}
\end{table*}

%% file: Tables/mteb_instructions.tex
\begin{table*}[ht]
\centering
\resizebox{2.1\columnwidth}{!}{%
\begin{tabular}{l|p{0.8\linewidth}}
\hline
\textbf{Task Name} & \textbf{Instruction} \\ \hline
AmazonCounterfactualClassification & Classify a given Amazon customer review text as either counterfactual or not-counterfactual \\
AmazonPolarityClassification & Classify Amazon reviews into positive or negative sentiment  \\
AmazonReviewsClassification & Classify the given Amazon review into its appropriate rating category  \\
Banking77Classification & Given a online banking query, find the corresponding intents  \\
EmotionClassification &  Classify the emotion expressed in the given Twitter message into one of the six emotions: anger, fear, joy, love, sadness, and surprise  \\
ImdbClassification & Classify the sentiment expressed in the given movie review text from the IMDB dataset  \\
MassiveIntentClassification & Given a user utterance as query, find the user intents  \\
MassiveScenarioClassification & Given a user utterance as query, find the user scenarios  \\
MTOPDomainClassification & Classify the intent domain of the given utterance in task-oriented conversation  \\
MTOPIntentClassification & Classify the intent of the given utterance in task-oriented conversation  \\
ToxicConversationsClassif. & Classify the given comments as either toxic or not toxic  \\
TweetSentimentClassification & Classify the sentiment of a given tweet as either positive, negative, or neutral  \\
ArxivClusteringP2P & Identify the main and secondary category of Arxiv papers based on the titles and abstracts  \\
ArxivClusteringS2S & Identify the main and secondary category of Arxiv papers based on the titles  \\
BiorxivClusteringP2P & Identify the main category of Biorxiv papers based on the titles and abstracts  \\
BiorxivClusteringS2S & Identify the main category of Biorxiv papers based on the titles  \\
MedrxivClusteringP2P & Identify the main category of Medrxiv papers based on the titles and abstracts  \\
MedrxivClusteringS2S & Identify the main category of Medrxiv papers based on the titles  \\
RedditClustering & Identify the topic or theme of Reddit posts based on the titles  \\
RedditClusteringP2P & Identify the topic or theme of Reddit posts based on the titles and posts  \\
StackExchangeClustering & Identify the topic or theme of StackExchange posts based on the titles  \\
StackExchangeClusteringP2P & Identify the topic or theme of StackExchange posts based on the given paragraphs  \\
TwentyNewsgroupsClustering & Identify the topic or theme of the given news articles  \\
SprintDuplicateQuestions & Retrieve duplicate questions from Sprint forum  \\
TwitterSemEval2015 & Retrieve tweets that are semantically similar to the given tweet  \\
TwitterURLCorpus & Retrieve tweets that are semantically similar to the given tweet  \\
AskUbuntuDupQuestions & Retrieve duplicate questions from AskUbuntu forum  \\
MindSmallReranking & Retrieve relevant news articles based on user browsing history  \\
SciDocsRR & Given a title of a scientific paper, retrieve the titles of other relevant papers  \\
StackOverflowDupQuestions & Retrieve duplicate questions from StackOverflow forum  \\
ArguAna & Given a claim, find documents that refute the claim  \\
ClimateFEVER & Given a claim about climate change, retrieve documents that support or refute the claim  \\
CQADupstackRetrieval &  Given a question, retrieve detailed question descriptions from Stackexchange that are duplicates to the given question  \\
DBPedia & Given a query, retrieve relevant entity descriptions from DBPedia  \\
FEVER & Given a claim, retrieve documents that support or refute the claim  \\
FiQA2018 & Given a financial question, retrieve user replies that best answer the question  \\
HotpotQA & Given a multi-hop question, retrieve documents that can help answer the question  \\
MSMARCO & Given a web search query, retrieve relevant passages that answer the query  \\
NFCorpus & Given a question, retrieve relevant documents that best answer the question  \\
NQ & Given a question, retrieve Wikipedia passages that answer the question  \\
QuoraRetrieval & Given a question, retrieve questions that are semantically equivalent to the given question  \\
SCIDOCS & Given a scientific paper title, retrieve paper abstracts that are cited by the given paper  \\
SciFact & Given a scientific claim, retrieve documents that support or refute the claim  \\
Touche2020 & Given a question, retrieve detailed and persuasive arguments that answer the question  \\
TRECCOVID & Given a query on COVID-19, retrieve documents that answer the query  \\
STS* & Retrieve semantically similar text.  \\
SummEval & Given a news summary, retrieve other semantically similar summaries  \\
\hline
\end{tabular}
}
\caption{The prompt instructions used for MTEB benchmark evaluation, following \citet{wang2024improving}. The "STS*" instruction applies to all STS tasks.}
\label{tab:mteb-instructions}
\end{table*}

%% file: Tables/full_mteb_result.tex
\begin{table*}[ht]
\centering
\resizebox{1.90\columnwidth}{!}{%
\begin{tabular}{l|ccc}
\hline
Dataset & Public Retrieval Data Only & Synthetic Data Only & Full Dataset \\ \hline
AmazonCounterfactualClassification & 80.1 & 78.7 & 79.2 \\
AmazonPolarityClassification & 94.0 & 94.4 & 95.9 \\
AmazonReviewsClassification & 51.8 & 54.1 & 55.8 \\
ArguAna & 60.2 & 51.4 & 61.3 \\
ArxivClusteringP2P & 48.1 & 50.7 & 50.3 \\
ArxivClusteringS2S & 46.0 & 47.2 & 46.9 \\
AskUbuntuDupQuestions & 64.2 & 66.3 & 66.1 \\
BIOSSES & 85.6 & 85.1 & 87.5 \\
Banking77Classification & 88.5 & 88.3 & 89.2 \\
BiorxivClusteringP2P & 37.7 & 43.8 & 42.7 \\
BiorxivClusteringS2S & 36.9 & 41.4 & 41.2 \\
CQADupstackRetrieval & 48.8 & 44.3 & 47.1 \\
ClimateFEVER & 35.4 & 26.0 & 37.8 \\
DBPedia & 51.5 & 45.8 & 52.3 \\
EmotionClassification & 51.2 & 53.4 & 51.9 \\
FEVER & 91.2 & 79.1 & 89.4 \\
FiQA2018 & 54.1 & 45.8 & 55.8 \\
HotpotQA & 77.6 & 57.9 & 75.9 \\
ImdbClassification & 90.3 & 93.4 & 94.2 \\
MSMARCO & 43.4 & 29.3 & 42.4 \\
MTOPDomainClassification & 96.3 & 95.7 & 96.6 \\
MTOPIntentClassification & 86.5 & 87.4 & 87.0 \\
MassiveIntentClassification & 80.1 & 80.6 & 80.3 \\
MassiveScenarioClassification & 82.1 & 81.8 & 82.4 \\
MedrxivClusteringP2P & 32.2 & 34.8 & 33.6 \\
MedrxivClusteringS2S & 32.5 & 35.4 & 34.8 \\
MindSmallReranking & 32.5 & 33.8 & 33.3 \\
NFCorpus & 39.4 & 37.9 & 38.5 \\
NQ & 65.9 & 57.7 & 66.9 \\
QuoraRetrieval & 89.5 & 86.0 & 89.1 \\
RedditClustering & 63.9 & 61.7 & 64.8 \\
RedditClusteringP2P & 66.8 & 64.1 & 67.3 \\
SCIDOCS & 22.0 & 23.7 & 22.7 \\
SICK-R & 83.5 & 80.4 & 83.8 \\
STS12 & 76.6 & 75.4 & 79.8 \\
STS13 & 86.8 & 86.6 & 88.3 \\
STS14 & 83.1 & 82.4 & 85.6 \\
STS15 & 88.5 & 88.6 & 91.3 \\
STS16 & 85.9 & 86.6 & 88.1 \\
STS17 & 91.7 & 87.0 & 91.9 \\
STS22 & 67.9 & 66.5 & 69.7 \\
STSBenchmark & 87.9 & 84.4 & 89.7 \\
SciDocsRR & 84.4 & 85.7 & 84.7 \\
SciFact & 78.6 & 74.1 & 76.4 \\
SprintDuplicateQuestions & 95.3 & 94.7 & 95.3 \\
StackExchangeClustering & 72.9 & 71.3 & 72.6 \\
StackExchangeClusteringP2P & 37.1 & 43.5 & 42.9 \\
StackOverflowDupQuestions & 54.1 & 54.7 & 55.0 \\
SummEval & 31.1 & 31.6 & 31.3 \\
TRECCOVID & 81.4 & 81.3 & 82.2 \\
Touche2020 & 23.3 & 26.6 & 24.6 \\
ToxicConversationsClassification & 66.9 & 69.8 & 68.8 \\
TweetSentimentExtractionClassification & 63.7 & 65.3 & 64.8 \\
TwentyNewsgroupsClustering & 53.4 & 53.2 & 53.8 \\
TwitterSemEval2015 & 81.1 & 77.5 & 81.5 \\
TwitterURLCorpus & 87.1 & 86.3 & 87.0 \\ \hline
Average & 65.9 & 64.5 & 67.0 \\
\hline
\end{tabular}
}
\caption{Complete MTEB evaluation results for each dataset. Detailed evaluation metrics and dataset information are available in \citet{muennighoff2023mteb}.}
\label{tab:full-mteb-result}
\end{table*}